\documentclass[lettersize,journal]{IEEEtran}
\usepackage{amsmath,amsfonts}
\usepackage{algorithmic}
\usepackage{algorithm}
\usepackage{array}
\usepackage[caption=false,font=normalsize,labelfont=sf,textfont=sf]{subfig}
\usepackage{textcomp}
\usepackage{stfloats}
\usepackage{url}
\usepackage{verbatim}
\usepackage{graphicx}
\usepackage{cite}
\usepackage{xcolor}
\usepackage{multirow}
\usepackage{booktabs}
\usepackage[colorlinks=false, linkcolor=blue, citecolor=blue, urlcolor=blue]{hyperref}
\begin{document}

\title{MagicPortrait: Temporally Consistent Face Reenactment with 3D Geometric Guidance}

\author{Mengting Wei, Yante Li, Tuomas Varanka, Yan Jiang, Guoying Zhao}

\markboth{Journal of \LaTeX\ Class Files,~Vol.~14, No.~8, August~2021}%
{Shell \MakeLowercase{\textit{et al.}}: A Sample Article Using IEEEtran.cls for IEEE Journals}

\IEEEpubid{0000--0000/00\$00.00~\copyright~2021 IEEE}

\maketitle

\begin{abstract}
In this study, we propose a method for video face reenactment that integrates a 3D face parametric model into a latent diffusion framework, aiming to improve shape consistency and motion control in existing video-based face generation approaches. Our approach employs the FLAME (Faces Learned with an Articulated Model and Expressions) model as the 3D face parametric representation, providing a unified framework for modeling face expressions and head pose. This not only enables precise extraction of motion features from driving videos, but also contributes to the faithful preservation of face shape and geometry. Specifically, we enhance the latent diffusion model with rich 3D expression and detailed pose information by incorporating depth maps, normal maps, and rendering maps derived from FLAME sequences. These maps serve as motion guidance and are encoded into the denoising UNet through a specifically designed Geometric Guidance Encoder (GGE). A multi-layer feature fusion module with integrated self-attention mechanisms is used to combine facial appearance and motion latent features within the spatial domain. By utilizing the 3D face parametric model as motion guidance, our method enables parametric alignment of face identity between the reference image and the motion captured from the driving video. Experimental results on benchmark datasets show that our method excels at generating high-quality face animations with precise expression and head pose variation modeling. In addition, it demonstrates strong generalization performance on out-of-domain images.
\end{abstract}


\section{Introduction}

Given a target face video, the task of face reenactment aims to animate a static face image using the video. In this process, it is essential to retain the source face's appearance and identity features, while accurately transferring the target facial pose to the generated face, which is commonly defined in neural face reenactment literature as head orientation and facial expressions \cite{ren2020deep,xu2021multi,yang2023deep,gan2023efficient}. Animating humans, animals, cartoons, and other general objects has garnered significant research interest in recent years. Among these, human face animation has been the most thoroughly investigated due to its wide-ranging applications in fields like social media, film production, and digital entertainment \cite{guan2023stylesync,prajwal2020lip,andlauer20213d,yu2023discrepancy}. Unlike conventional graphics-based methods, the availability of large-scale data facilitates the creation of cost-effective, data-driven animation frameworks.

\begin{figure*}[htbp]
    \centering
    \includegraphics[width=0.95\textwidth]{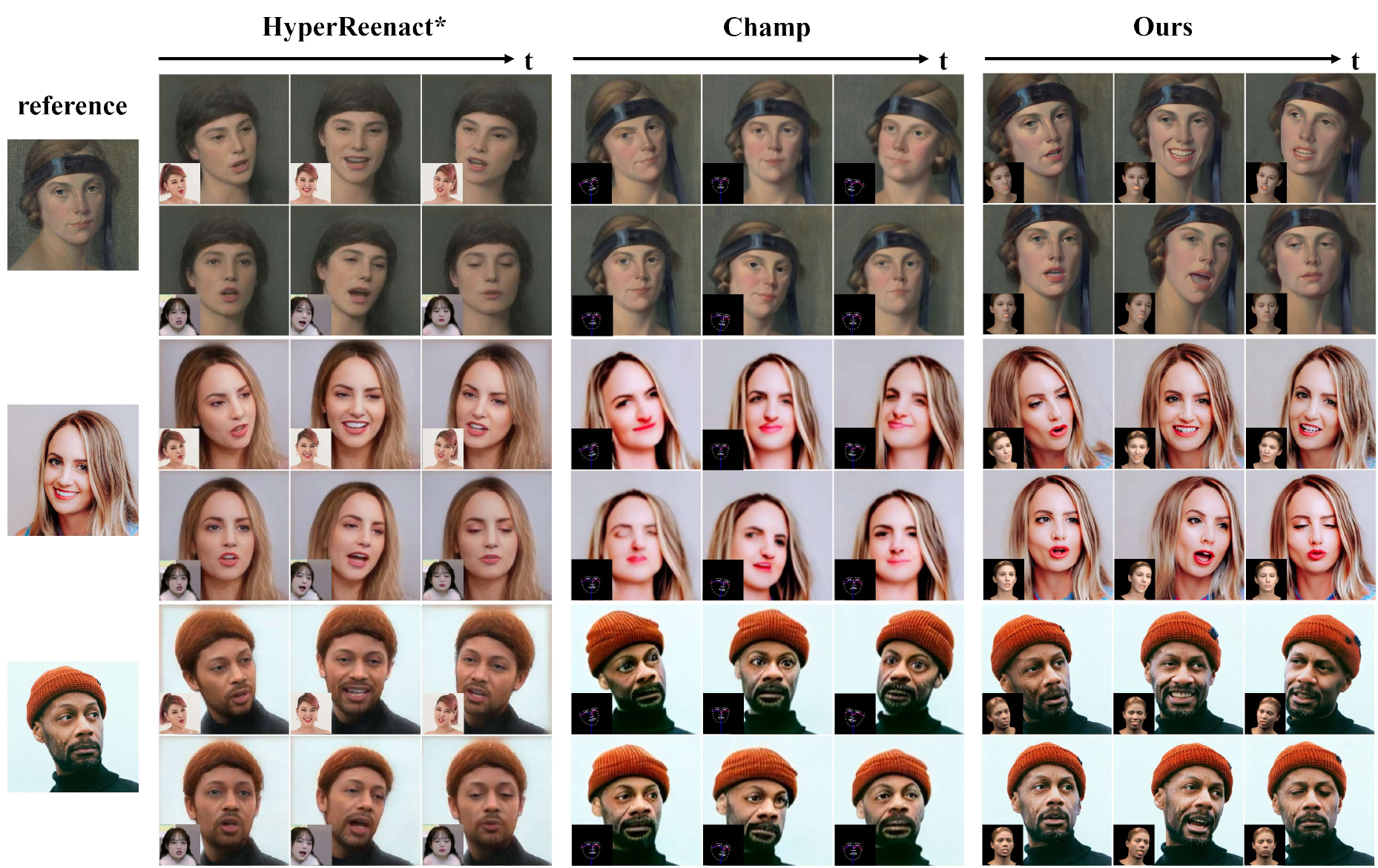}  
    \caption{Given a sequence of motion signals, MagicPortrait (ours) is capable of generating temporally coherent animations for reference face images. In contrast, state-of-the-art approaches struggle to maintain appearance details or transfer precise facial expressions. The motion sequence is shown in the corner for reference. *Note: HyperReenact utilizes raw video frames directly as the driving signal. Both Champ and our method utilize conditions extracted from the same driving video as used in HyperReenact.}
    \label{fig:demo}
\end{figure*}

A central challenge in face reenactment lies in learning disentangled representations of identity, facial pose and expressions, while also capturing a wide range of variations in facial attributes such as geometry, background, lighting, etc \cite{hempel2024toward,andlauer20213d,he2019attgan}. Current data-driven approaches for face reenactment can be broadly divided into two main categories according to the underlying generative models: GAN-based \cite{xu2021relightgan,xu2021multi,jiang2018graph,yu2022cmos} and diffusion-based frameworks \cite{li2025diffusion,ho2020denoising,rombach2022high}. Such approaches can produce high-quality and visually convincing synthetic images and offer a certain level of controllability in the generation process. Although recent generative models have been widely applied to face reenactment tasks, the problem remains highly challenging due to the significant diversity in facial attributes. Inadequate disentanglement often causes identity leakage from the driving face to the source face. To address these challenges, some works have turned to pre-trained StyleGAN2 \cite{karras2020analyzing} models to leverage their photo-realistic generation capabilities and inherent disentanglement properties \cite{bounareli2022finding,oorloff2023robust,bounareli2023hyperreenact,yin2022styleheat}. However, a key limitation lies in the difficulty of accurately encoding real images into the StyleGAN2 latent space, which frequently results in inaccurate reconstructions and suboptimal reenactment quality. As shown in Fig. \ref{fig:demo}, HyperReenact \cite{bounareli2023hyperreenact}, a StyleGAN2-based method, struggles to preserve critical appearance details such as hairstyles and hats, which are essential for maintaining the perceptual fidelity of the reenacted faces. Moreover, most GAN-based methods operate on images and process long videos frame by frame, subsequently stacking the results along the temporal axis \cite{hong2022depth,pang2023dpe,yang2022face2face}. Such approaches often overlook temporal consistency, leading to noticeable flickering artifacts.

\IEEEpubidadjcol

On the other hand, diffusion-based approaches condition the generation process on both appearance and motion by incorporating the reference image along with dynamic cues \cite{zeng2023face,du2023dae,stypulkowski2024diffused}. These methods enable the direct synthesis of animation videos as well as high identity consistency by leveraging the generative power of latent diffusion models combined with conditional guidance. In particular, human body animations usually condition on a sequence of poses or skeletons \cite{ren2020deep,zhang2022unsupervised,hu2024animate} to animate a reference image and synthesize controllable animation videos. However, these methods often lack accurate facial expression transfer, as the facial landmarks in the skeleton representation are insufficient to precisely capture detailed expression changes. As shown the Champ \cite{zhu2024champ} model in Fig. \ref{fig:demo}, which uses landmarks to describe face, fails to accurately capture and transfer facial expression dynamics due to the sparsity of facial landmarks. Moreover, when there are significant geometric differences between the source and driving faces, such as variations in camera distance or facial shape, using landmarks or contours from the driving video can lead to shape inconsistencies in the reenacted face relative to the source. As shown in the third example of Fig. \ref{fig:demo}, the image synthesized by Champ \cite{zhu2024champ} appears noticeably broader than the reference, indicating a distortion in body shape preservation. In contrast, 3D parametric face models offer the ability to disentangle face shape from expressions and poses, which implies that the source face shape can be effectively combined with the expression and pose dynamics of the driving input. \cite{jiang20183d,lou2021real}. This facilitates a more convenient and direct transfer of expressions and poses during inference. Nevertheless, even with 3D parametric models, many approaches continue to employ 3D-projected landmarks for motion guidance \cite{wei2024aniportrait,yang2022face2face}, thereby limiting the accuracy of expression transfer.  Therefore, based on a more accurate and appropriate 3D representation as motion guidance, the aim of this paper is to further optimize facial expression guidance and face shape preservation mechanisms for high-quality of video face reenactment.

In this study, we argue that combining a reference image with motion guidance, which is typically provided by sequential skeleton data or landmarks information, presents certain limitations in terms of both facial expression alignment and shape preservation. As a further development, we recommend the use of a 3D parametric face model FLAME \cite{li2017learning} to model the 3D geometry of the reference face and extract motion cues from the driving video. Firstly, unlike keypoints-based approaches that mainly focus on head pose, the FLAME model provides a unified framework that jointly captures face shape, expression and pose variations within a compact, low-dimensional parameter space. Consequently, the FLAME model not only provides expression and pose information but also delivers structural guidance related to surface deformation, spatial interactions (e.g., occlusions), contours, and various face-dependent features. Secondly, the parametric formulation of the FLAME model enables the alignment of geometric structures between the FLAME mesh reconstructed from the identity image and the motion representations obtained from the driving video. This allows for the refinement of FLAME-based motion sequences, thereby improving the conditioning of motion and geometric shape in latent diffusion models. Due to FLAME’s ability to generalize across diverse face shapes, it effectively accommodates significant differences in face shape between the reference image and the driving video.

By integrating the FLAME model as a guidance mechanism for both facial expression and head pose, our approach is organized into four core components within the latent diffusion framework: 1) The FLAME-based motion sequences extracted from the driving video are projected into the image space, producing depth maps, normal maps and rendering maps that collectively encode key 3D information. The depth map is essential for representing the spatial structure of the face, while the normal map captures its surface orientation. The rendering maps enhance the precision of guidance information for intricate facial expressions. 2) In the inference stage, Structured Face Alignment facilitates the integration of the source facial geometry with the expression and pose from the driving video, ensuring that the reenacted face remains consistent with the source’s structural features. 3) A Geometric Guidance Encoder (GGE) is designed to process these maps, facilitating the integration of their information into the feature space of the denoising UNet. 4) During the integration of guidance maps through feature encoding, self-attention is applied to enable the network to focus on salient regions of the source image across different feature layers. This hierarchical semantic fusion significantly strengthens the model's ability to perceive and preserve source image details with greater accuracy. Finally, conditioning the latent video diffusion model on these multi-level feature representations enables accurate generation of animated images with faithful pose and expression consistency.

We validate the effectiveness of our approach through extensive experiments on the widely used CelebV-HQ video dataset, where it significantly enhances the quality of face reenactment. In addition, we benchmark our method against state-of-the-art techniques on out-of-domain data comprising videos and images from diverse face image styles, highlighting its strong generalization ability.

\section{Related Work}

\subsection{Face Reenactment}

Most current face reenactment techniques employ controllable GANs that are conditioned on various facial cues or attributes, such as facial landmarks or keypoints. A subset of these methods utilizes facial landmarks \cite{zakharov2019few,zakharov2020fast}, typically extracted using off-the-shelf landmark detection models like \cite{bulat2017far}, to guide the generation process. However, these landmark-based approaches often suffer from identity leakage in cross-subject reenactment scenarios, since landmarks inherently preserve the facial structure of the target individual. To address this issue, several works \cite{xu2021relightgan,cao2021unifacegan,wu2023poce} explore the use of 3D Morphable Models (3DMMs) \cite{blanz2023morphable}, which offer a disentangled representation of facial identity and pose. By decoupling facial shape from pose, 3DMM-based methods are more effective at maintaining identity consistency across different subjects. In parallel, warping-based approaches attempt to learn intermediate keypoints that capture facial motion between the source and target faces, offering another route for motion transfer and reenactment. However, warping-based methods often result in noticeable artifacts or unnatural distortions in the generated reenacted faces.

A recent stream of research \cite{hou2023semi,pernuvs2023maskfacegan} explores the photo-realistic image synthesis capabilities of pre-trained StyleGAN2 models \cite{karras2020analyzing}. Bounareli et al. \cite{bounareli2022finding} examine the $\mathcal{W}$ latent space of StyleGAN2 to identify directions that control facial pose modification. Similarly, Oorloff et al. \cite{oorloff2023robust} utilize a combination of the $\mathcal{W}$ space and style space $\mathcal{S}$ to enable effective face editing. HyperReenact \cite{bounareli2023hyperreenact} employs a hypernetwork \cite{ha2016hypernetworks} to dynamically modify the weights of the StyleGAN2 generator, enabling simultaneous refinement of the source identity and transfer of the target’s facial pose. While the method effectively captures and conveys the target’s head orientation and expressions, it often results in overly smooth reenacted images, leading to the loss of fine-grained visual details from the source, such as glasses, background elements, or hairstyles. Moreover, most of these GAN-based approaches handle video frames in isolation, ignoring the temporal dynamics inherent in animation sequences, which inevitably results in flickering artifacts in the generated videos.

\subsection{Diffusion-based Portrait Animation}

Diffusion models \cite{zhou2025recurrent,liu2024towards} generate target data samples by progressively denoising from an initial Gaussian distribution. In \cite{rombach2022high}, Latent Diffusion Models (LDMs) were introduced to perform both training and inference within a compressed latent space, significantly improving computational efficiency. Leveraging their versatility, LDMs have been widely adopted in a range of applications, including full-body dance generation \cite{hu2024animate,xu2024magicanimate}, audio-driven portrait animation \cite{wei2024aniportrait,qi2023difftalker}, and video-based portrait reenactment \cite{zeng2023face}. Unlike image generation, video generation poses additional challenges such as maintaining temporal coherence and consistency across frames. To address this, recent approaches \cite{ho2020denoising,xie2024x} often extend pre-trained diffusion-based image generators by incorporating temporal modeling layers, aiming for joint spatial-temporal representation learning. Several studies \cite{xie2024x,wei2024aniportrait} have adopted mutual self-attention mechanisms and incorporated temporal attention modules which is similar to the architecture used in AnimateAnyone \cite{hu2024animate} to enhance image fidelity and improve the preservation of appearance. AniPortrait \cite{wei2024aniportrait} utilizes spatial cues specifically keypoints as intermediate representations to guide facial motion. In contrast, X-Portrait \cite{xie2024x} bypasses such intermediate motion representations and directly drives the portrait animation using the original video input. To enable cross-identity training, it adopts an implicit keypoint-based approach as introduced in \cite{wang2021one}. Many human body animation methods also utilize keypoints to transfer facial motion. However, we argue that keypoints alone are insufficient to accurately capture the subtle variations in facial expressions. To address this limitation, we employ more expressive 3D renderings as the conditioning input for facial expressions.

\section{Methodology}

In this section, we first introduce the background of the FLAME model \cite{li2017learning} and the latent diffusion model. We then describe how FLAME is employed as a condition for reenactment, followed by a detailed explanation of the design of the Geometric Guidance Encoder (GGE). Finally, we present the overall model architecture along with the training and inference pipeline.

\subsection{Reenactment Motion Guidance}
\label{subsec:mg}

\textbf{Structured Face Alignment.} A central challenge in face video generation is animating a reference image with a motion sequence while faithfully preserving the subject’s appearance and shape. Existing landmark-based approaches typically rely on sparse keypoints as motion guidance, which limits their ability to capture the full diversity of face shapes. By leveraging a parametric face model, our method facilitates effective alignment of both shape and motion between the reference identity and the motion sequence. As shown in the top right panel of Fig. \ref{fig:framework}, given a FLAME model $F_{id}$ fitted to the reference image $I_{id}$, and a FLAME sequence $F_{d}^{1: N}$ obtained from the driving video $I_{1:N}$, our objective is to transfer the identity parameters $\boldsymbol{\beta}_{id}$ of $F_{id}$ to the pose sequence $\boldsymbol{\theta}$ of $F_{d}^{1: N}$ and the expression sequence $\boldsymbol{\psi}$ of $F_{d}^{1: N}$. For each frame $i \in[1, N]$, the corresponding aligned FLAME model is constructed as:

\begin{equation}
F_{\mathrm{infer}}^i=\operatorname{FLAME}\left(\boldsymbol{\beta}_{\mathrm{id}}, \boldsymbol{\psi}_{d}^i,\boldsymbol{\theta}_{d}^i\right)
\end{equation}

The rendered conditions from $F_{\mathrm{infer}}^{1:N}$ are subsequently employed to guide video generation from $I_{id}$, , enabling pixel-wise face shape alignment and enhancing the mapping of face appearance in the synthesized animation.


\subsection{Geometric Motion Conditioning}

At this stage, we have completed shape-level alignment between the FLAME model reconstructed from the identity image and the FLAME sequence derived from the driving video, using a structured shape alignment approach. Based on the aligned FLAME sequence, we then render depth maps, normal maps, and rendering maps to serve as rich structural and appearance cues. By incorporating latent feature embeddings and the self-attention mechanism described below, we are able to apply spatial weighting across multi-layer representations of face and motion. This process enables the generation of a multi-layer semantic fusion that serves as effective motion guidance.

\textbf{Geometric Guidance Encoder.} ControlNet \cite{zhang2023adding} is commonly adopted in style transfer tasks to guide generated images with additional conditional inputs. However, incorporating multiple guidance signals into ControlNet can lead to a significant computational overhead, making it impractical for efficient deployment. Motivated by the recent advances in \cite{hu2024animate}, we propose a geometric guidance encoder tailored to process multilevel conditioning inputs. This design enables the concurrent extraction of informative features from the guidance signals while adapting a pre-trained denoising UNet. The encoder is composed of a sequence of lightweight subnetworks. We design a dedicated guidance module for each conditioning input to extract its corresponding features. Initially, each guidance input is processed through a series of convolutional layers to capture low-level representations. Given the multilevel nature of the guidance conditions, which reflect diverse aspects of the face, a self-attention module is integrated following the convolutional layers. This module enhances the ability to capture and associate relevant semantic information across the multiple layers of guidance inputs. To maintain the structure and performance of the pre-trained denoising UNet, we employ a convolutional layer with zero-initialized weights as the output layer, ensuring the controlled extraction of features from each guidance condition. A detailed architecture of the guidance encoder is illustrated in the GGE of Fig. \ref{fig:framework}.

The feature embeddings extracted from each guidance condition are combined within the guidance encoder through an element-wise summation, producing the final guidance representation $\boldsymbol{C}$, which is formulated as:

\begin{equation}
\boldsymbol{C}=\sum_{m=1}^M \mathcal{G}^m\left(\cdot, \theta^m\right).
\end{equation}
\noindent Here, $M=3$ denotes the total number of incorporated guidance conditions, $m$ represents the index corresponding to the motion guidance, $\theta$ refers to the input guidance image (depth, normal and rendering), and $\mathcal{G}$ is the Geometric Guidance Encoder. The resulting guidance feature is then fused with the noisy latent representation before being passed into the denoising fusion module.

\subsection{Model Architecture}

\textbf{Model Structure.} In this section, we will detail the full workflow of our method illustrated in Fig. \ref{fig:framework}. The core of our approach is a video diffusion model enhanced by motion guidance based on 3D face parametric models. Concretely, the FLAME model is employed to extract a continuous head sequence from the driving motion input, yielding guidance that captures both 2D projections and 3D geometric information. For efficient integration of the guidance signals, a sequence of GGEs is designed to process the motion guidance. These latent embeddings are individually enhanced using self-attention and subsequently combined through a multi-layer feature fusion module. To maintain temporal and appearance consistency in the generated video, we introduce two essential components: the AppearanceNet and the temporal alignment module. The source image is encoded into the AppearanceNet, which plays a crucial role in preserving coherence between the appearance and background of characters across the generated video and the identity image. The architectures of AppearanceNet and the Denoising UNet are identical, and both are initialized with the UNet parameters from Stable Diffusion.

\textbf{Multi-layer Feature Fusion.} Our design is inspired by the architectures of AnimateAnyone \cite{hu2024animate} and MagicAnimate \cite{xu2024magicanimate}, which have demonstrated that feature fusion within the spatial-attention layers of two identical UNets can effectively preserve the visual details of the source image. Specifically, let $\boldsymbol{h}_A$ and $\boldsymbol{h}_D$ denote the hidden states before each Transformer block of AppearanceNet and the Denoising UNet, respectively. We perform element-wise addition of $\boldsymbol{h}_A$ and $\boldsymbol{h}_D$ to obtain the fused hidden state $\boldsymbol{h}_{fu}$, which is then passed to the spatial attention layer of the subsequent Transformer block of the Denoising UNet, shown on the far right of Fig. \ref{fig:framework}. For the cross-attention, although many existing works replace the original text embeddings in the cross-attention layers of Stable Diffusion with CLIP-encoded \cite{radford2021learning} features to enhance alignment with the generated images \cite{zhu2024champ,hu2024animate}, we observe that this approach can lead to slower convergence and a noticeable degradation in final generation quality. Consequently, we utilize the text embedding corresponding to the prompt "\textit{a close-up of a person}" as the consistent conditioning input in the cross-attention layers for all samples.

\textbf{Temporal Alignment Module.} Drawing inspiration from AnimateDiff \cite{guo2023animatediff}, which extends the capabilities of a text-to-image model to generate videos from text prompts, we inserts a series of temporal modules into the Denoising UNet to perform attention-based modeling across video frames. This approach is designed to improve the temporal coherence and visual consistency of the generated video.

\textbf{Training and Inference Protocol.} The training process is divided into two distinct phases. The primary goal of the first phase is to synthesize a high-quality animated image conditioned on the guidance information derived from the target frame. In this phase, the model is trained using only images, with temporal modules deactivated.  During this stage, the parameters of the VAE encoder and decoder, along with the text encoder, are kept frozen. In contrast, the GGEs, Denoising UNet, and AppearanceNet are updated through gradient-based optimization. At the beginning of this stage, one frame is randomly selected from a face video to act as the identity image, while another frame from the same video is chosen as the target. The guidance conditions extracted from the target frame is then passed through the GGEs. In the second phase of training, the motion module is released to enhance the temporal consistency and smoothness of the generated video. A video clip containing 12 frames is selected and used as input. During this stage, the GGEs, Denoising UNet, and AppearanceNet, which have already been trained in the first phase, remain frozen and are not updated. The training objective can be derived as:

\begin{equation}
\mathcal{L}=\mathbb{E}_{\mathbf{z}_t, \boldsymbol{C}, \mathbf{z}_{id}, t}\left[\left\|\epsilon-\boldsymbol{\phi}\left(\mathbf{z}_t, t, \boldsymbol{C}, \mathbf{z}_{id} \right)\right\|_2^2\right],
\end{equation}
\noindent where $\epsilon$ is the noise added to the sample, $\boldsymbol{\phi}$ is the Denoising UNet, and $\mathbf{z}_{id}$ is the latent features passed from AppearanceNet. The text condition is omitted in this formulation since it remains unchanged during the entire process.

During inference, a given identity face image is animated by aligning motion sequences obtained from either in-the-wild videos or synthetically generated ones. To facilitate this alignment, we apply the Structured Face Alignment introduced in Sec. \ref{subsec:mg} to match the motion sequence with the reconstructed FLAME model of the identity image at the pixel level, serving as the foundation for animation. We follow the temporal aggregation strategy \cite{tseng2023edge} to concatenate multiple segments of 12 frames, which enables the generation of longer video outputs with improved temporal continuity.

\section{Experiments}

\subsection{Implementation Details}

\noindent \textbf{Dataset.} Our model is trained on a randomly sampled subset of the CelebV-HQ dataset \cite{zhu2022celebv}. CelebV-HQ is a large-scale and high-quality video dataset featuring diverse subjects and comprehensive facial attribute annotations. It consists of 35,666 video clips with a minimum resolution of 512×512, covering 15,653 unique identities. From this dataset, we randomly sampled 1,200 video clips to construct the training set for our experiments.

\noindent \textbf{Implementation.} Our experiments were conducted using 8 AMD MI250 GPUs to ensure sufficient computational resources. The training pipeline consists of two distinct phases. In the initial phase, individual video frames are preprocessed by resizing them to a standardized resolution of 512×512 pixels. The first stage of training was conducted for 45,000 steps using a batch size of 32. Both AppearanceNet and Denoising UNet are inititalized from SD 1.5\footnote{https://huggingface.co/stable-diffusion-v1-5/stable-diffusion-v1-5}. In the second stage, we focused on optimizing the temporal layers by training with 12-frame video sequences for 50,000 steps, employing a reduced batch size of 1 on two NVIDIA A100 GPUs to better capture temporal coherence. We adopt the DDIMScheduler \cite{song2020denoising} as our sampling scheduler. A consistent learning rate of 1e-5 was applied throughout both stages. The temporal layers are initialized using AnimateDiff \cite{guo2023animatediff}. During inference, to ensure smooth continuity across longer sequences, we applied a temporal aggregation strategy \cite{tseng2023edge} that enables seamless merging of outputs from separate batches, resulting in the generation of extended video sequences.

\begin{figure*}[htbp]
    \centering
    \includegraphics[width=0.95\textwidth]{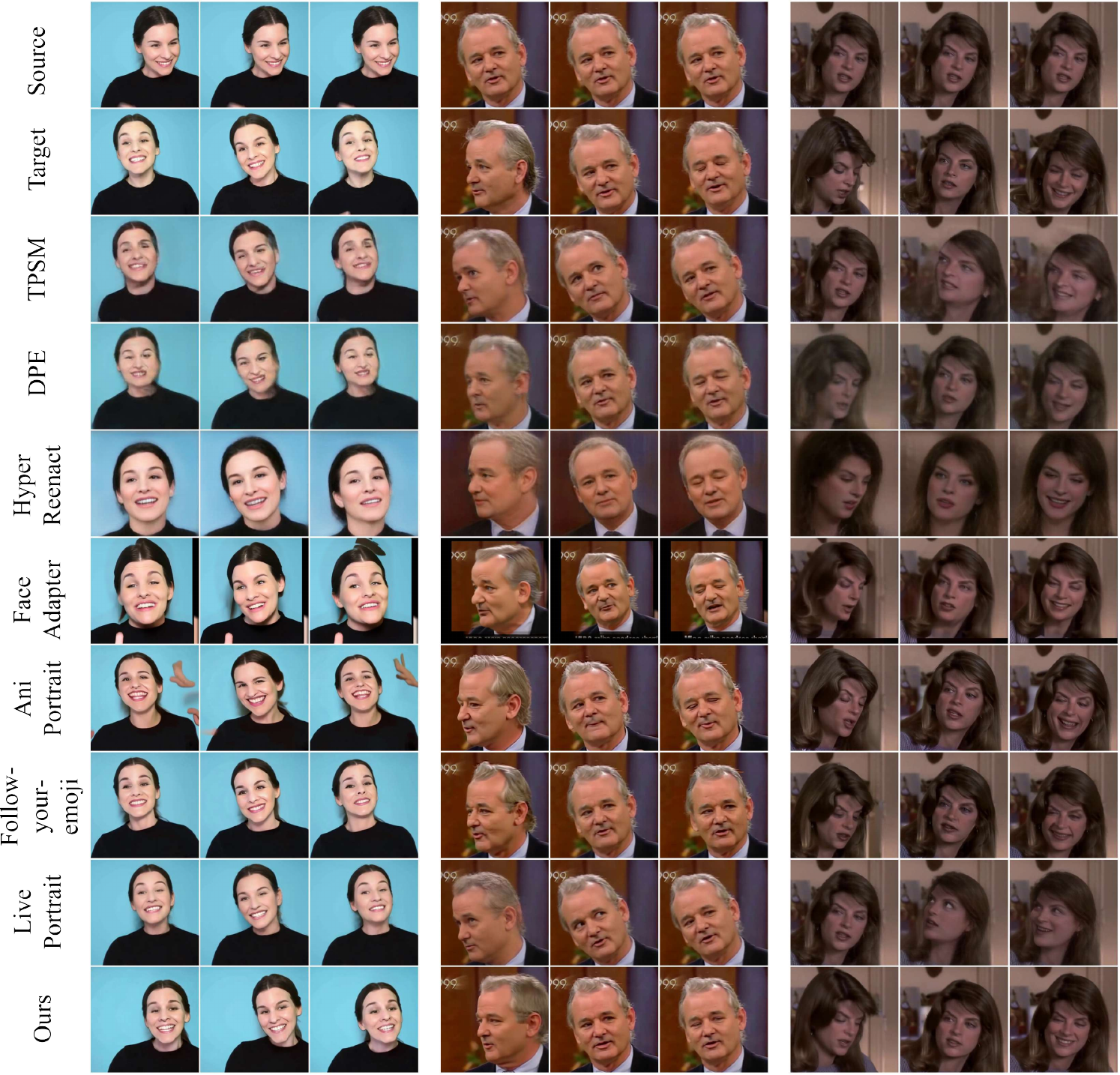}  
    \caption{Qualitative comparisons on self-reenactment on samples from CelebV-HQ dataset.}
    \label{fig:self}
\end{figure*}

\subsection{Comparisons to state-of-the-art}

In this section, we present both quantitative and qualitative evaluations on self- and cross-subject reenactment tasks. We adopt eight evaluation metrics in accordance with previous literature. To assess the reconstruction quality of the compared methods, we measure identity similarity (CSIM) via the cosine similarity of features from a face recognition model \footnote{https://github.com/ageitgey/face\_recognition}. To evaluate the visual quality of individual frames, we employ commonly used metric LPIPS \cite{zhang2018unreasonable} which collectively reflects perceptual similarity. Although SSIM \cite{wang2004image} and PSNR \cite{hore2010image} are also commonly used metrics in animation tasks, they may not be entirely appropriate for evaluating face reenactment. In face reenactment, different methods often involve variations in input preprocessing and output resolution, which can cause shifts in the facial location within the generated images, even when the pose and expression are accurately transfered, as shown by the results from HyperReenact \cite{bounareli2023hyperreenact} and LivePortrait \cite{guo2024liveportrait} in Fig. \ref{fig:self}. Consequently, relying on pixel-wise differences for evaluation may lead to unfair comparisons. To address this, we adopt the FID \cite{heusel2017gans} metric for a more robust and fair assessment. Furthermore, to evaluate the accuracy of head pose transfer, we compute the Average Pose Distance (APD) derived by calculating the mean L1 distance of the pose parameters extracted by SMIRK \cite{retsinas20243d} between the animated and the driving images. We also compute root mean square error between the target and reenacted pose coefficients (P-RMSE) estimated by Deep3DFace \cite{deng2019accurate}. For facial expressions transfer, we use Average Expression Distance (AED) derived by calculating the mean L1 distance of the expression features obtained from the pre-trained EMOCA \cite{danvevcek2022emoca} model and root mean square error between the target and reenacted expression coefficients (E-RMSE) estimated by Deep3DFace. To evaluate the fidelity of the generated videos, we adopt Fréchet Video Distance (FVD) \cite{unterthiner2018towards} as an independent measure.

\begin{table*}[htbp]
 \small
  \centering
  \caption{Quantitative comparisons of self-reenactment on the test images from CelebV-HQ dataset. The best and second best results are reported in \textbf{bold} and \underline{underline}, respectively.}
  \begin{tabular}{lcccccccc}
    \toprule
    \multirow{2}{*}{Method~~}&\multicolumn{3}{c}{\textbf{Image}} &\multicolumn{2}{c}{\textbf{Pose}} &\multicolumn{2}{c}{\textbf{Exp}} &\multicolumn{1}{c}{\textbf{Video}}\\
    \cmidrule(r){2-4} \cmidrule(r){5-6} \cmidrule(r){7-8} \cmidrule(r){9-9} & CSIM\(\uparrow\)~~ & LPIPS\(\downarrow\)~~  &FID\(\downarrow\)~~ & APD\(\downarrow\)~~ &P-RMSE\(\downarrow\)~~ & AED\(\downarrow\)~~ & E-RMSE\(\downarrow\)~~ & FVD\(\downarrow\)\\
    \midrule
    TPSM \cite{zhao2022thin}~~          &0.8518~~ &0.157~~ & 77.41~~ &0.087~~ &0.170~~ &0.080~~  &0.232~~  &206.08\\
    DPE \cite{pang2023dpe}~~          &\underline{0.8876}~~ &\underline{0.132}~~ & 60.35~~ &0.039~~ &0.076~~ &0.137~~  &0.426~~  &232.08\\
    HyperReenact \cite{bounareli2023hyperreenact}~~ &0.7687~~ &0.337~~ & 69.21~~ &0.047~~ &0.082~~ &0.083~~ &~~0.251~~ ~ &\underline{168.82}\\
    LivePortrait \cite{guo2024liveportrait}~~ &0.8605~~ &0.239~~ &58.22~~ &0.104~~ &0.188~~ &\underline{0.046}~~ &\underline{0.167}~~  &\textbf{156.29} \\
    \midrule
    FaceAdapter \cite{han2024face}~~  &0.7091~~ &0.349~~ &90.43~~ &\underline{0.029}~~ &\textbf{0.061}~~ &0.129~~ &0.422~~  &485.51 \\
    Aniportrait \cite{wei2024aniportrait}~~  &0.8517~~ &0.165~~ &72.11~~ &0.033~~ &0.067~~ &0.115~~ &0.394~~ &226.49 \\
    Follow-your-emoji \cite{ma2024follow}~~ &0.8753~~ &0.149~~ &\underline{56.80}~~  &0.052~~ &0.106~~ &0.052~~ &0.186~~ &211.43 \\
    \midrule
    \textbf{Ours} ~~ &{\bf 0.9102}~~ &{\bf 0.125}~~ &{\bf 53.68}~~ &\bf{0.027}~~ &\underline{0.064}~~ &\textbf{0.036}~~ &\bf{0.146}~~  &179.13 \\
    \bottomrule
  \end{tabular}
  \label{tab:self}
\end{table*}

\noindent \textbf{Self-reenactment task.} For the self-reenactment task, we compare our approach with 7 state-of-the-art methods. In particular, we include comparisons with 4 controllable GAN-based methods trained from scratch, namely TPSM \cite{zhao2022thin}, DPE \cite{pang2023dpe}, HyperReenact \cite{bounareli2023hyperreenact} and LivePortrait \cite{guo2024liveportrait}. We also provide comparisons with 3 SD-based methods, namely FaceAdapter \cite{han2024face}, AniPortrait \cite{wei2024aniportrait} and Follow-your-emoji \cite{ma2024follow}. For all baseline methods, we utilize the officially released pre-trained models that are publicly available. We randomly sampled 40 different identities which is distinct from the training set in the CelebV-HQ dataset to form part of our test set.

As shown in the comparisons of Tab. \ref{tab:self}, as measured by APD and P-RMSE, our approach achieves comparable performance to DPE, Follow-your-emoji and FaceAdapter in terms of head pose transfer, which exhibits a negligible error when compared to the target.  Furthermore, our method surpasses all baselines on CSIM, LPIPS, and FID metrics, indicating its ability to generate high-quality images and meanwhile to maintain the identity. With respect to facial expression transfer, our method remains competitive when compared to state-of-the-art video generation models including LivePortrait, Follow-your-emoji and Aniportrait. On the other hand, from Fig. \ref{fig:self}, we observe that our method preserves the relative spatial position of the original face within the frame, while seamlessly modifying the head pose and facial expressions. For instance, the upper body of the leftmost person consistently remains in the bottom-right corner of the image across all frames. This design choice is meaningful, as significant differences in the absolute position of the target face compared to the source would require the model to perform extensive inpainting. Such discrepancies are prone to introducing artifacts, particularly in the cross-subject reenactment setting. Although our method shows slightly inferior scores in video quality metric FVD, some competing approaches achieve lower values by keeping portions of the background static or overly smooth throughout the sequence. This strategy can lead to noticeable distortions when there are significant head movements, as observed from the rightmost person of LivePortrait.


\begin{table*}[htbp]
 \small
  \centering
  \caption{Ablation study on different motion guidance. "lmks" denotes a scenario where only the landmark map is utilized as the motion guidance.}
  \begin{tabular}{lcccccccc}
    \toprule
    \multirow{2}{*}{Method~~}&\multicolumn{3}{c}{\textbf{Image}} &\multicolumn{2}{c}{\textbf{Pose}} &\multicolumn{2}{c}{\textbf{Exp}} &\multicolumn{1}{c}{\textbf{Video}}\\
    \cmidrule(r){2-4} \cmidrule(r){5-6} \cmidrule(r){7-8} \cmidrule(r){9-9} & CSIM\(\uparrow\)~~ & LPIPS\(\downarrow\)~~  &FID\(\downarrow\)~~ & APD\(\downarrow\)~~ &P-RMSE\(\downarrow\)~~ & AED\(\downarrow\)~~ & E-RMSE\(\downarrow\)~~ & FVD\(\downarrow\)\\
    \midrule
    only lmks~~          &0.8691~~ &0.143~~ & 62.25~~ &0.083~~ &0.101~~ &0.087~~  &0.240~~ &305.17\\
    lmks+depth+normal~~  &\underline{0.8942}~~ &0.133~~ &57.40~~ &0.049~~ &0.073~~ &0.047~~ &\underline{0.165}~~ &226.86 \\
    Ours (depth+normal)~~  &0.8909~~ &\underline{0.127}~~ &\textbf{53.44}~~ &\bf{0.024}~~ &\bf{0.064}~~ &0.055~~ &0.180~~ &\underline{200.77} \\
    Ours (only rendering)~~ &0.8733~~ &0.134~~ & 60.09~~ &0.030~~ &\underline{0.069}~~ &\underline{0.041}~~ &\underline{0.165}~~ &221.15\\
    \midrule
    Ours (full configuration)~~ &{\bf 0.9102}~~ &{\bf 0.125}~~ &\underline{53.68}~~ &\underline{0.027}~~ &\bf{0.064}~~ &\textbf{0.036}~~ &\bf{0.146}~~  &\bf{179.13} \\

    \bottomrule
  \end{tabular}
  \label{tab:abl1}
\end{table*}

\noindent \textbf{Cross-identity reenactment task.} On the cross-subject reenactment task, the source and target faces are from different subjects, which poses greater implementation difficulty. Therefore, to validate the effectiveness of our method, we perform evaluations using out-of-domain images. We collected 20 high-resolution identity images from the public image-sharing platform \footnote{https://unsplash.com} to further augment the test set. These 20 identity images exhibit diverse visual styles, such as oil painting, comic, crayon drawing, anime style, and naturalistic backgrounds. In addition, we substitute TPSM \cite{zhao2022thin}, DPE \cite{pang2023dpe} and HyperReenact \cite{bounareli2023hyperreenact} with two alternative approaches built upon Stable Diffusion, i.e., Champ \cite{zhu2024champ} and StableAnimator \cite{tu2024stableanimator}, both of which employ facial landmarks as conditional inputs to guide facial expression transfer. Qualitative and quantitative comparisons are presented in Fig. \ref{fig:cross} and Tab. \ref{tab:cross}. 

Observations from the first and third rows of Fig. \ref{fig:cross} indicate that our approach maintains robust generalization performance, even when applied to facial images that substantially deviate from real human facial appearances. Consistently, our approach obtains the highest CSIM value in Tab. \ref{tab:cross}, which reflects its effectiveness in maintaining identity fidelity when applied to out-of-domain inputs. On the other hand, our approach outperforms landmark-based methods (e.g., Champ, StableAnimator, Aniportrait, Follow-your-emoji) in terms of facial expression transfer, highlighting its enhanced ability to capture and replicate subtle expression variations. Furthermore, our approach demonstrates improved capability in maintaining the original head scale from the identity image, as reflected in the qualitative comparisons with FaceAdapter, follow-your-emoji and LivePortrait. These results suggest that incorporating 3D conditions provides effective guidance for modeling facial geometry.

\begin{figure}[htbp]
    \centering
    \includegraphics[width=0.49\textwidth]{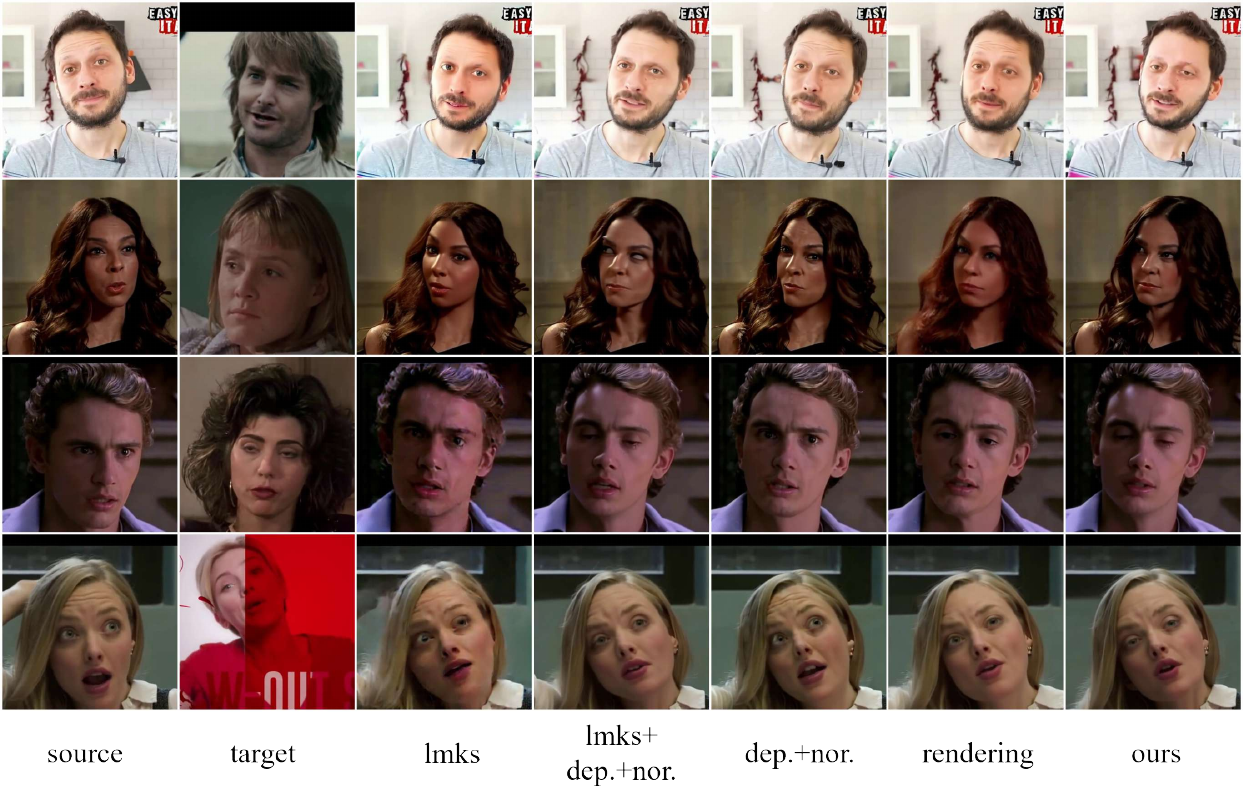}  
    \caption{Ablation analysis on different motion conditions. Lmks, dep. and nor. stand for landmarks, depth and normal, respectively. Zoom in to compare the details.}
    \label{fig:abl1}
\end{figure}

\subsection{Ablation Study}

\begin{table*}[htbp]
 \small
  \centering
  \caption{Ablation study on the architecture of Geometric Guidance Encoder.}
  \begin{tabular}{lcccccccc}
    \toprule
    \multirow{2}{*}{Method~~}&\multicolumn{3}{c}{\textbf{Image}} &\multicolumn{2}{c}{\textbf{Pose}} &\multicolumn{2}{c}{\textbf{Exp}} &\multicolumn{1}{c}{\textbf{Video}}\\
    \cmidrule(r){2-4} \cmidrule(r){5-6} \cmidrule(r){7-8} \cmidrule(r){9-9} & CSIM\(\uparrow\)~~ & LPIPS\(\downarrow\)~~  &FID\(\downarrow\)~~ & APD\(\downarrow\)~~ &P-RMSE\(\downarrow\)~~ & AED\(\downarrow\)~~ & E-RMSE\(\downarrow\)~~ & FVD\(\downarrow\)\\
    \midrule
    CLIP enc.~~          &0.8914~~ &0.157~~ & 54.34~~ &0.170~~ &0.237~~ &0.268~~  &0.677~~  &301.15\\
    w/o. self-attn~~  &0.8996~~ &0.140~~ &57.79~~ &0.037~~ &0.071~~ &0.044~~ &0.162~~  &241.66 \\
    w/. self-attn~~  &{\bf 0.9102}~~ &{\bf 0.125}~~ &{\bf 53.68}~~ &\bf{0.027}~~  &\bf{0.064}~~ &\textbf{0.036}~~ &\bf{0.146}~~  &\bf{179.13} \\

    \bottomrule
  \end{tabular}
  \label{tab:abl2}
\end{table*}

\begin{figure}[htbp]
    \centering
    \includegraphics[width=0.49\textwidth]{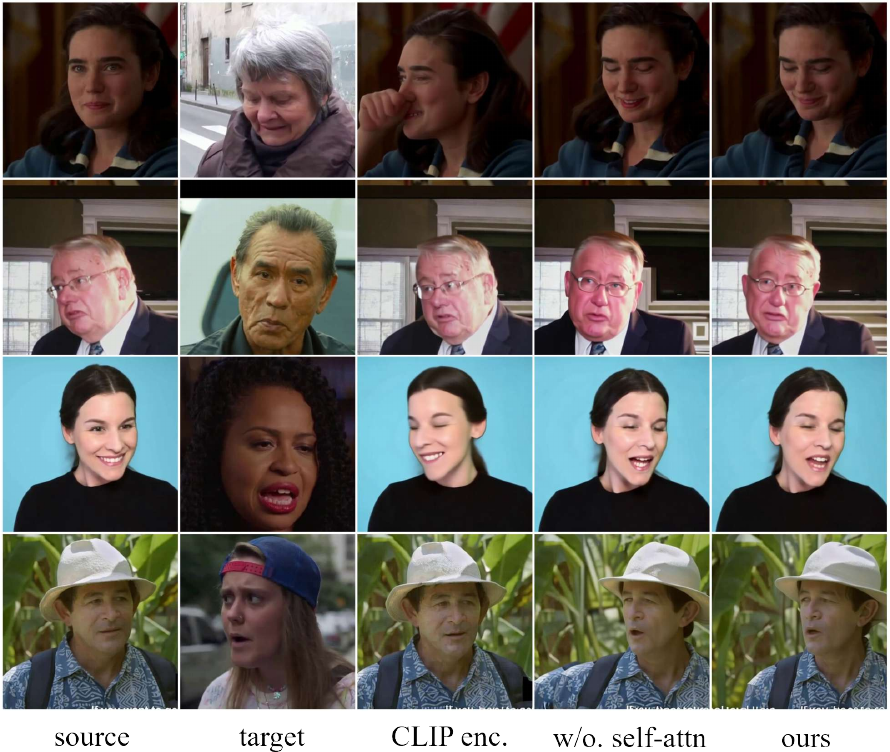}  
    \caption{Ablation analysis on different architecture of Geometric Guidance Encoder. w/o. indicate the guidance without self-attention.}
    \label{fig:abl2}
\end{figure}


\begin{figure}[htbp]
    \centering
    \includegraphics[width=0.49\textwidth]{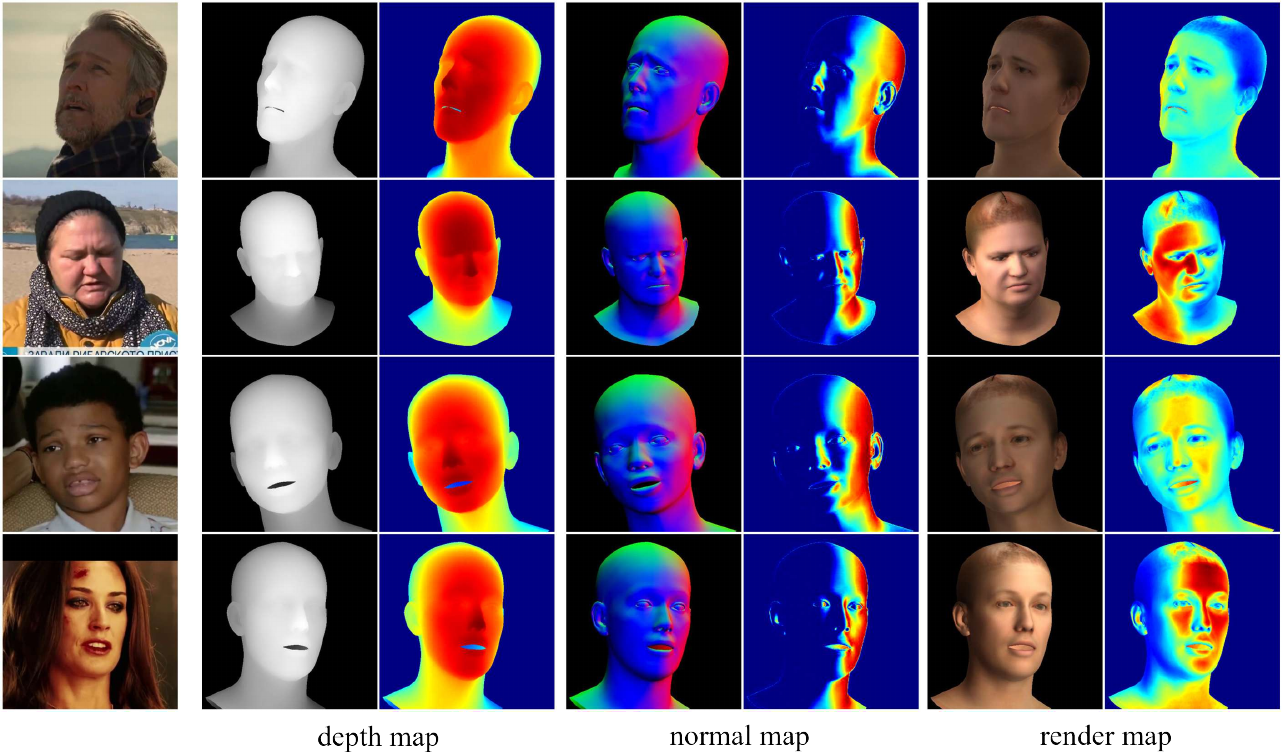}  
    \caption{The motion conditions alongside their corresponding self-attention maps. The left column of each group showcases the depth map, normal map, and rendering map generated from the associated FLAME sequences. The right column displays the resulting self-attention outputs obtained from the guidance encoder.}
    \label{fig:vis}
\end{figure}

\noindent \textbf{FLAME motion guidance.} We evaluate the impact of different FLAME motion conditions by progressively incorporating and replacing motion guidance. The experiments are conducted under five different settings: 1) using only landmark-based images as motion guidance; 2) combining depth, normal, and landmark-based images as motion guidance; 3) using only depth and normal maps as geometric conditions; 4) using only renderings as motion conditions; and 5) using the full configuration of our original method. All settings are conducted using the data for the self-reenactment task. As presented in Tab. \ref{tab:abl1}, the quantitative results indicate that the complete configuration of our proposed method significantly surpasses all ablated variants in terms of image quality, visual fidelity, and temporal consistency. The decrease in the AED and E-RMSE metrics from setting 3) to 2) to 5) highlights the importance of the rendering condition for effectively transferring facial expressions. FLAME yields greater improvements in CSIM (0.9102 vs. 0.8733) and LPIPS (0.125 vs. 0.134) compared to the only rendering-based approach, indicating its superior capability in maintaining accurate face shape alignment. Moreover, a comparison between configurations 3) and 5) reveals that rendering achieves lower AED and E-RMSE scores, suggesting that it facilitates more accurate expression transfer. In addition, Fig. \ref{fig:abl1} visually demonstrates the effectiveness of the various motion conditions.

\noindent \textbf{Geometric Guidance Encoder.} We conduct comparisons of different GGE architectures. Specifically, we first replace the GGE with a CLIP image encoder \cite{radford2021learning} and Linear layers. Additionally, we retain the GGE structure but substitute the internal self-attention modules with InflatedConv3D layers. Tab. \ref{tab:abl2} summarizes the results of the ablation study performed within the guidance module. The findings demonstrate that incorporating guidance attention consistently enhances performance, as reflected by improvements in all evaluation metrics. Meanwhile, as illustrated in Fig. \ref{fig:abl2}, the CLIP encoder demonstrates limited ability in guiding target pose and expression. This may be attributed to the fact that image features extracted by CLIP primarily capture high-level semantics, which are less effective compared to pixel-aligned conditions for conveying fine-grained spatial and structural information. By comparing the results with and without self-attention, we observe that the absence of self-attention tends to produce incorrect details in the generated images. For instance, the window in the background of the second row and the shirt of the person in the fourth row are inaccurately rendered without self-attention mechanisms.


\noindent \textbf{Structured Face Alignment.} We conduct an ablation study on the alignment of face shape coefficients between the FLAME model of the appearance individual and the FLAME sequence derived from the driving video. To emphasize the impact of face alignment, we select a subject with a distinctive face shape as the reference and use a typical face video as the motion input. As shown in Fig. \ref{fig:abl3}, compared to other approaches, the results in the last column obtained through structured face alignment show superior consistency in face shape and figure preservation with respect to the appearance image.

\noindent \textbf{Visualization.} Fig. \ref{fig:vis} presents the self-attention maps of depth, normal and rendering feature embeddings after training. The analysis reveals clear distinctions among different guidance types: the depth condition primarily captures the geometric contours of the face; the normal condition highlights surface orientation and fine-grained structural details; and the render condition focuses on facial regions and local appearance variations, which are critical for preserving facial expression details. These observations confirm that the multi-level guidance effectively decomposes and captures complementary aspects of face shape and motion, contributing to the generation of more realistic and temporally consistent face animations.

\subsection{Limitations and Future Works}

\noindent \textbf{Limitations.} In real-world scenarios, facial images are often captured with camera motion, which can lead to background changes across frames. If the training videos contain a large number of such samples, this may result in minor temporal artifacts, such as background flickering in the synthesized outputs. Despite the high quality of the CelebV-HQ dataset, this problem persists in a significant portion of its video sequences.

\noindent \textbf{Future Work.} Our approach can be combined with human body animation methods to improve the overall quality of full-body animation, as many existing works predominantly condition on skeletons or semantic maps and do not adequately capture facial expressions.

\section{Conclusion}

This paper presents a new methodology for face reenactment by integrating the FLAME 3D parametric model with latent diffusion frameworks, aiming to improve both face alignment and motion control. By utilizing the unified representation of head pose and face shape provided by FLAME, and incorporating depth and normal maps the proposed approach further enhances the capacity to capture realistic motion dynamics and anatomical structures compared to existing methods. The integration of rendering guidance and self-attention mechanisms for feature fusion further enhances the animation pipeline, enabling the generation of dynamic visual content that closely mirrors face anatomy and motion. Extensive experiments across multiple datasets validate the effectiveness of the proposed method in producing high-quality face animations, highlighting its potential to drive advancements in digital content creation where detailed and realistic human face depictions are essential.

\bibliographystyle{IEEEtran}
\bibliography{tip_references}
\vfill

\end{document}